\documentclass[runningheads]{llncs}
\usepackage{graphicx}
\usepackage{amsmath,amssymb} 
\usepackage{color}
\makeatletter
\newcommand{\printfnsymbol}[1]{%
  \textsuperscript{\@fnsymbol{#1}}%
}
\makeatother

\begin{document}

\title{Robust Human Matting via Semantic Guidance}
\titlerunning{Robust Human Matting via Semantic Guidance}
%
%

\author{Xiangguang Chen\thanks{These authors contributed equally to this work.}\inst{1} \and
Ye Zhu\printfnsymbol{1}\inst{2} \and
Yu Li\thanks{Corresponding Author.}\inst{3}\and
Bingtao Fu \inst{1}\and
Lei Sun \inst{1} \and
Ying Shan \inst{2} \and
Shan Liu \inst{1}
}

\authorrunning{X. Chen et al.}

\institute{Platform Technologies, Tencent Online Video \and
ARC Lab, Tencent PCG \\
\email{\{seanxgchen,samuelzhu\}@tencent.com} \and
International Digital Economy Academy (IDEA) \\
\email{liyu@idea.edu.cn}}

\maketitle

\begin{abstract}
Automatic human matting is highly desired for many real applications. We investigate recent human matting methods and show that common bad cases happen when semantic human segmentation fails. This indicates that semantic understanding is crucial for robust human matting. From this, we develop a fast yet accurate human matting framework, named Semantic Guided Human Matting (\textbf{SGHM}). It builds on a semantic human segmentation network and introduces a light-weight matting module with only marginal computational cost. Unlike previous works, our framework is data efficient, which requires a small amount of matting ground-truth to learn to estimate high quality object mattes. Our experiments show that trained with merely 200 matting images, our method can generalize well to real-world datasets, and outperform recent methods on multiple benchmarks, while remaining efficient. Considering the unbearable labeling cost of matting data and widely available segmentation data, our method becomes a practical and effective solution for the task of human matting. Source code is available at \url{https://github.com/cxgincsu/SemanticGuidedHumanMatting.}

\end{abstract}

\section{Introduction}
Human matting aims to predict an alpha matte to extract human foreground from an input image or video, which has many important applications in visual processing. To achieve that, a green screen is often required for studio solutions. However, a green screen is not always available in many real scenarios, such as daily video conferencing and background replacement effects shot with mobile devices. Therefore, human matting methods without a green screen are highly desired. Many previous works use an additional trimap for matting, which indicates three kinds of regions in an image, namely foreground, background, and unknown. However, it requires careful manual annotation to obtain a trimap. Background matting approaches~\cite{bgmv2,bgmv1} are recently proposed which use a pre-recorded background image as a prior. Though decent results are obtained, it only can handle cases with a static background and a fixed camera pose.

\begin{figure}[ht]
\centering
\includegraphics[width=\linewidth]{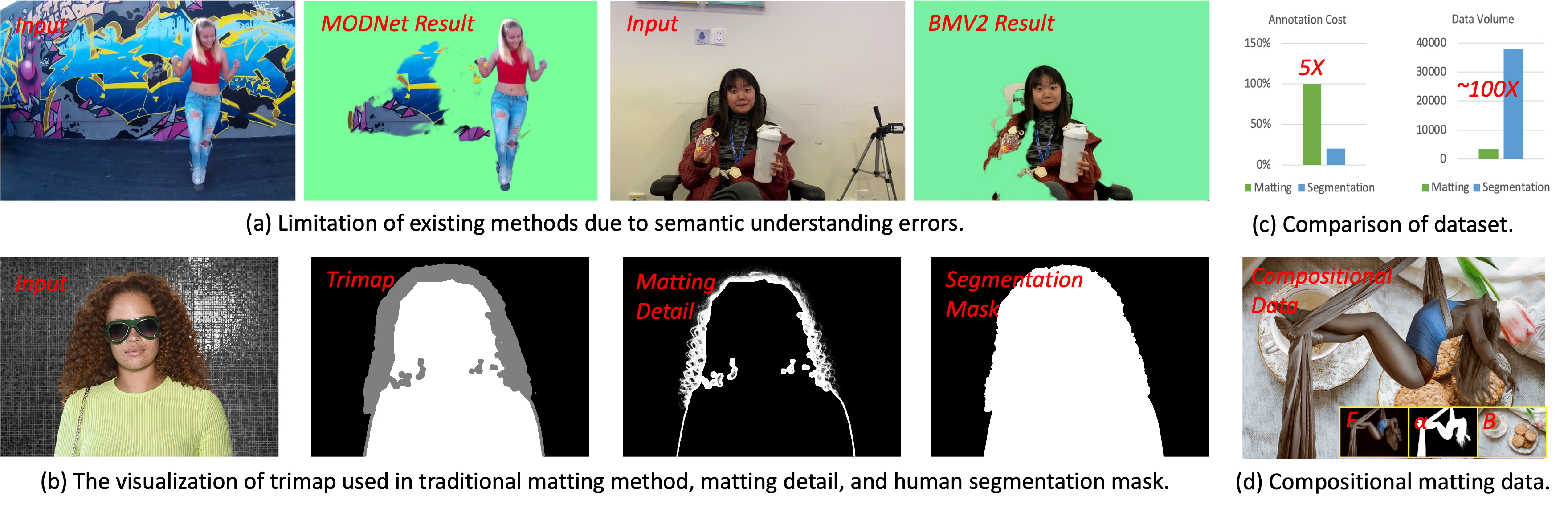}
\caption{Limitation of existing method and motivations of this work. (a) Common failure cases of latest works~\cite{bgmv2,MODNet} happen when semantic understanding fails. (b) Traditional matting methods rely on the input of a trimap. Since matting details are located around the human mask boundaries~\cite{yu2021mask}, the coarse segmentation mask can also be leveraged as a prior in matting.  (c) Currently, segmentation data is much easier to annotate, and the amount of publicly available data is much larger than that of matting data.  (d) Compositing foreground with different backgrounds can enlarge the matting dataset size but it has a domain gap as it looks unreal~\cite{li2022bridging}.} 
\label{fig:motivation}
\end{figure}

Many recent works focus on developing methods towards automatic human matting. Some early attempts~\cite{chen2018semantic,shen2016deep} try to generate pseudo trimap as a first step and predict a matte from the trimap. Due to limited training data, these methods cannot generalize well to real-world examples~\cite{bgmv1}. Another drawback of these methods is that they cannot run in real-time which is required for many applications, such as background replacement in live video conferencing. The recent work MODNet~\cite{MODNet} proposes a fast and fully automatic portrait matting method. RVM~\cite{lin2022robust} is another recent work which leverages temporal information in a video to improve robustness and stability. 

In this work, we aim to develop a robust, accurate, and fast method for automatic human matting, which shares the same goal as MODNet~\cite{MODNet} and RVM~\cite{lin2022robust}. We investigate the failure cases of existing automatic methods on real-world examples and observe that these failure cases are often due to inaccurate semantic understanding. As shown in Fig.~\ref{fig:motivation} (a), parts of the background are wrongly predicted as foreground or part of human body are wrongly segmented. This indicates a weak semantic understanding ability of these state-of-the-art (SOTA) methods. In order to enhance their ability of semantic understanding while keeping fine-grained details of matting, we seek to utilize semantic segmentation task to guide matting process. There are three reasons behind this motivation. 1) Segmentation mask determines the overall accuracy of foreground and background predictions, and fine-grained structures only appear around the mask. This indicates that a semantic human mask can replace a trimap (Fig.~\ref{fig:motivation} (b)) and be used as a prior condition for matting~\cite{yu2021mask}. 2) The labeling of high-quality matting requires skillful annotators and is very time-consuming. For that, the amount of available training data for matting is quite limited (at the order of hundreds and thousands) compare to segmentation task, which require only simple line drawings around boundaries. As a matter of fact, there are many human segmentation datasets at a scale that is two or more magnitude larger (Fig.~\ref{fig:motivation} (c)). A larger amount of data is of great significance to the generalization ability on real-world images. 3) Synthetic datasets created by compositing images (Fig.~\ref{fig:motivation} (d)) are also used in training matting models, but they have a clear limitation due to the drastic domain gap between synthetic and real-world images. This prevents the trained models from generalizing to real-world examples. The work~\cite{li2022bridging} analyzes the domain gap issue systematically. Our approach does not suffer from this issue by using less of such data.

Based on the above analysis, We propose a multi-stage framework to predict semantic segmentation mask and matting alpha successively. A segmentation sub-network is first employed for the task of segmentation, and then it is reused to guide the matting process to focus on the surrounding area of the segmentation mask. To achieve real-time efficiency as well as better performance, we let the two tasks share the encoder part of the model, which has been proved superior to separated encoders in~\cite{li2022bridging}. By this design, our matting module successfully handled many challenging cases. In summary, our network consists of a shared encoder, a segmentation decoder and a matting decoder, and the segmentation decoder feeds useful intermediate information to the matting decoder. In training, a two-stage pipeline is proposed. Firstly, the encoder and the segmentation decoder are trained with publicly available segmentation datasets. With these data, our segmentation sub-network is trained to predict robust human masks. Secondly, 269 matting images are employed to train the matting decoder. To comprehensively evaluate the performance of matting methods, we adopt 5 benchmarks to carry out qualitative and quantitative comparison. One of them is our self-collected dataset from complex scenarios, such as diverse background, multiple human, body accessories, and low light. Our method outperforms all other methods across all benchmarks. 

We summarize our contribution as follows:

\begin{enumerate}
    \item We develop a robust, accurate and efficient human matting framework, which utilizes shared encoder for both segmentation and matting. It gives our method the ability to use powerful semantic understanding to guide matting process meanwhile help to reduce computation.
    \item The proposed framework can make fully use of coarse mask training data and reduce matting reliance on high-quality and large number of annotations. With only about 200 matting images, our method is able to produce high quality alpha details.
    \item Extensive experiments show our method achieves the state-of-the-art results on multiple benchmarks.
\end{enumerate}

\section{Related Work}
In this section, we review matting with auxiliary input and automatic matting, which are related to our work. We also review segmentation as segmentation provides the rough mask of human region.

\textbf{Matting with auxiliary input.} Early methods are mostly optimization or filter based which require an additional trimap as  input~\cite{aksoy2017designing,chen2013knn,chuang2001bayesian,gastal2010shared,levin2007closed,levin2008spectral,sun2004poisson,chen2004grayscale,pham2009real,park2010convex,sindeev2012alpha}. Deep learning is introduced in trimap-based matting methods in \cite{deepimagematting,fbamatting,liu2021tripartite} that use a deep network for trimap-based matting. These trimap-based methods are often general to different matting target objects but it requires the user to provide trimap annotations. Background mattings~\cite{bgmv1,bgmv2} are recently proposed to replace the trimap input with  a pre-recorded background image as a prior condition. Although background matting can generate decent results on static background, it cannot be applied to camera moving circumstances. Recently proposed mask-guided method~\cite{yu2021mask} achieves SOTA results once a coarse is provided. In their work, mask is generated from manual annotation or segmentation output, which greatly limits the convenience of use. Our goal is to incorporate the mask generation into the matting process, so as to realize fully automatic matting and still keeping real-time running.

\textbf{Automatic matting.} Fully automatic matting without any additional input has been pursued \cite{yang2022unified,dai2022boosting,chen2022pp}. Methods in \cite{qiao2020attention,zhang2019late} studies class agnostic matting but cannot generalize well.
Some methods like \cite{chen2018semantic,shen2016deep,zhu2017fast,Sun_2022_CVPR,xing2022composite} dedicate to human matting. In this direction, the latest MODNet~\cite{MODNet} aims at fast portrait matting and RVM~\cite{lin2022robust} is towards robust human matting using temporal information. For MODNet, it performs well in the portrait image, but easily fails in full body image. Recent work P3M-Net~\cite{li2021privacy} proposes a dual decoder to do human matting, which is similar to us. But there are several significant differences: 1) P3M-Net use segmentation decoder to generate a pseudo trimap while our segmentation predicts real mask. P3M-Net predicts alpha details only on trimap unknown region. This setting tends to output false matting results when trimap is wrongly predicted. Our matting decoder treats mask as guidance and regresses alpha at the whole image. Under this setting, the matting decoder is given an opportunity to correct semantic errors. 2) Our segmentation decoder and matting decoder are trained at two separate stages. At the segmentation training stage, the segmentation decoder is strongly supervised by a large dataset. As a result, the segmentation decoder predicts more robust results than the weakly supervised result in P3M-Net. 3) Another advantage of our model is it is data-efficient in that we only use a very small amount of high-precision data to train the matting decoder.

\textbf{Segmentation.} Semantic segmentation assigns a semantic class label to every pixel in the scene. Its difference with matting is that it predicts a hard binary mask that belongs to either foreground or background and cannot generate fine details and transparent value as in matte. So directly applying segmentation mask to image and video composition will generate hard boundary at the foreground object, leaving noticeable artifacts when replacing the backgrounds. However, segmentation can provide strong semantic cues of the object location which facilitate our matting task. Many deep learning-based semantic segmentation are fully convolutional and some effective modules like Atrous Spatial Pyramid Pooling (ASPP)~\cite{deeplabv3} are proposed. We follow them in our segmentation network design.

\section{Method}
Given a color image $I$, the matting task can be formulated as follows:
\begin{equation}
    I = \alpha F+(1-\alpha)B,~\alpha~\in~[0,1],
\end{equation}
where $F, B$ are foreground and background, and $\alpha$ is the alpha matte denoting where is foreground part located. For image matting problem, we should predict the alpha matte from the input color image, which is a hard and ill-posed task. As mentioned earlier, existing methods rely on additional auxiliary inputs like trimap or pre-captured background. Automatic method like RVM is not robust against semantic error. Based on this, we try to design a framework to better leverage the semantic prior from segmentation, but produce fine detail and transparent matte values. A straightforward way is to rely on a semantic segmentation mask and generate the matting results using a new matting network. This setup is developed and demonstrated in mask-guided (MG) matting~\cite{yu2021mask}. The two-step setup treats segmentation and matting as two separate tasks and has a few drawbacks. First of all, the matting network only uses the predicted segmentation map and ignores the rich semantic features. Second, using a separate matting network will  extract features again from image and introduce additional computation, which slows down the speed noticeably on high resolution. 

\begin{figure*}[htb]
	\centering
		\includegraphics[width=\linewidth]{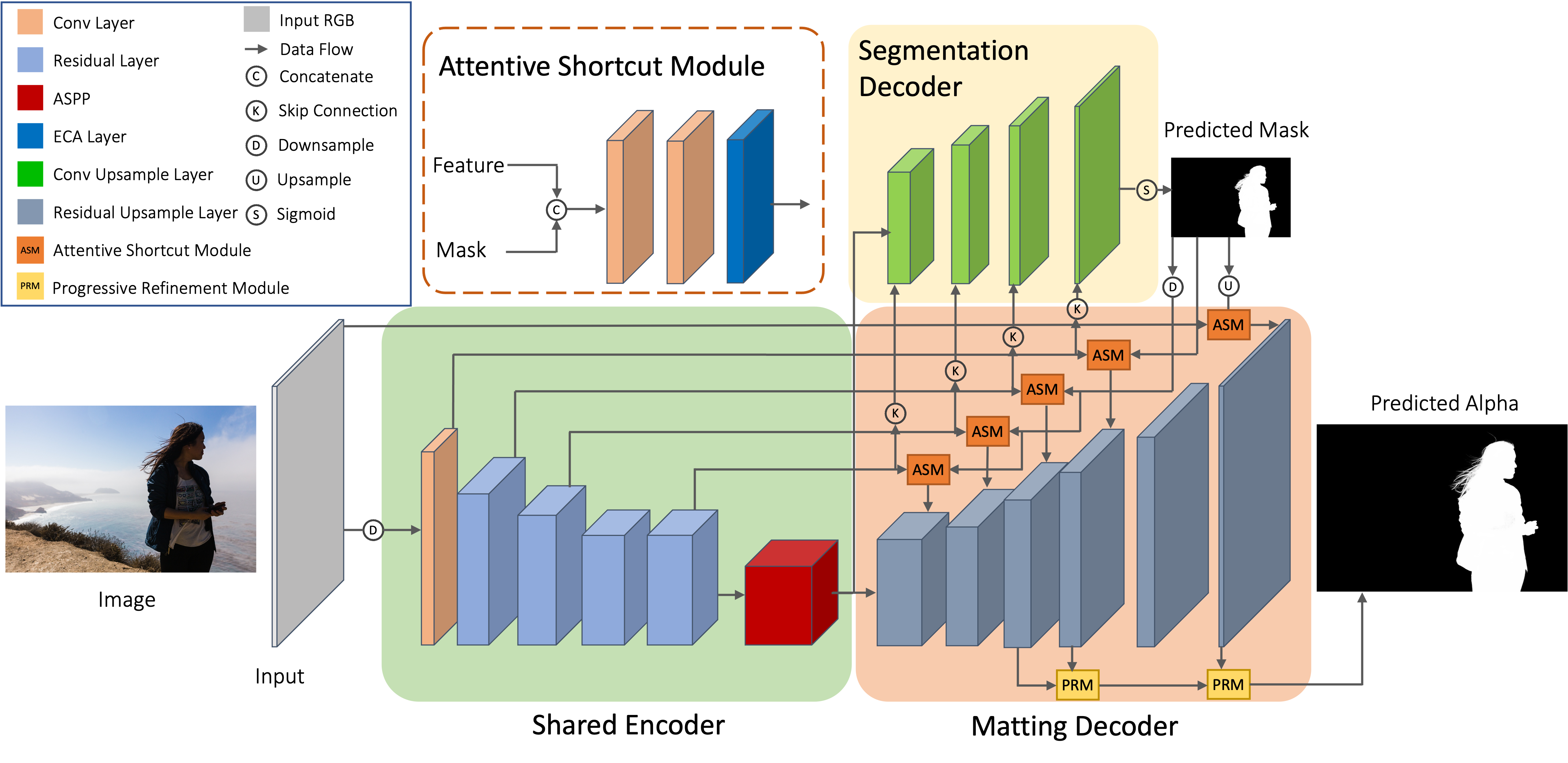}
	\caption{{The network structure of our SGHM. High-resolution image is first downsampled for the shared encoder, then the segmentation decoder is used to generate a coarse semantic mask prediction. We propose an Attentive Shortcut Module(ASM) to adaptively fuse shared features and masks. Finally, the matting decoder refines the unknown area of human margin and predicts the alpha matte.}}\label{fig:pipeline}
\end{figure*}

Based on above analysis, we propose a new human matting method named Semantic Guided Human Matting (SGHM), which uses a segmentation network to guide human matting. Specifically, we share the encoder between segmentation and matting task. Thus, matting task can learn accurate semantic understanding from reusing the rich semantic features in encoder and focus on predicting alpha details in matting decoder. 

As shown in Fig.~\ref{fig:pipeline}, our SGHM consists of a shared encoder to extract image features, a segmentation decoder to predict image segmentation mask, and a matting decoder with Progressive Refinement Module (PRM)~\cite{yu2021mask} to predict a high-resolution matting result. We propose to use an Attentive Shortcut Module (ASM) to combine the features from encoder and mask from segmentation decoder for matting decoder.

\subsection{Shared Encoder}
As mentioned above, we propose to improve matting results by using semantic human segmentation features. So we make the segmentation and matting tasks share an encoder. More specifically, we first train the encoder and segmentation decoder as segmentation model, and then fix the parameters of the encoder and train the matting decoder with segmentation features extracted from encoder. We adopt ResNet50~\cite{resnet} as feature extraction backbone followed by a ASPP module~\cite{deeplabv3} for shared encoder, which extracted features at $\frac{1}{4}$, $\frac{1}{8}$, $\frac{1}{16}$, $\frac{1}{32}$, $\frac{1}{64}$ scale for two decoders with an input image at $\frac{1}{4}$ scale, which can be denoted as $F_0$, $F_1$, $F_2$, $F_3$, $F_4$.

\subsection{Segmentation Decoder}
Our segmentation decoder is a light-weight and efficient module, which contains $4$ convolution layers and $4$ up-sample layers. For each convolution layer, it can be defined as:
\begin{equation}
X_i = {\rm{Conv}( Concat( Upsample(} X_{i+1} ), F_i ), i = 3,2,1,0,
\end{equation} 
where $F_i$ is the feature from shared encoder and $X_i$ is the output feature of convolution layer. In particular, $X_4 = F_4$ is the direct input of segmentation decoder. Following each convolution layer, a batch normalization layer and a ReLU layer are attached except the last one. Finally, we obtain the output segmentation mask $S$. We denote our segmentation branch as SGHM-S in reporting the results later.

\subsection{Matting Decoder}
Our matting decoder inputs the segmentation features and segmentation mask of different scales, outputs the matting results at $1,\frac{1}{4},\frac{1}{8}$ scale. Firstly, we use ASM module to combine the features from encoder and segmentation mask. Then we sequentially process the features of different scales by several upsample blocks. We predict matting results at $1,\frac{1}{4},\frac{1}{8}$ scale by output modules. Finally, we adopt PRM module to produce the final high-resolution matting result based on the matting results at three output scales.

\noindent \textbf{Attentive Shortcut Module.} 
Our model proposed to use semantic segmentation to improve human matting by sharing  encoder of segmentation and matting. In addition to features from shared encoder , we also feed the segmentation mask of different scales as input of matting decoder. For matting decoder, how to fuse the features and mask from segmentation is of vital importance. One direct way is to concatenate these two inputs for further processing. We propose to use ASM to fuse these two inputs. With the help of ASM, we can get more adaptive features for matting decoder. Specifically, the ASM contains two convolution layers, two SpectralNorm layers ~\cite{miyato2018spectral} and an efficient channel attention layer~\cite{wang2020eca}. Channel attention can produce an adaptive feature by calculating a channel-wise weight vector corresponding to input feature.

\noindent \textbf{Upsample Block.} Upsample block process input features sequentially from $\frac{1}{64}$ scale to the original scale. First, it element-wisely adds the feature of the current scale and the feature of the previous scale upsampled by residual blocks from $\frac{1}{64}$ scale to $\frac{1}{2}$ scale. Then, for $\frac{1}{2}$ scale and $1$ scale, we replace the residual blocks with a single transposed convolution layer with batch normalization and ReLU for efficiency. 

\noindent \textbf{Output Block.} We predict matting result at $1, \frac{1}{4},\frac{1}{8}$ scale. For each output scale, we attach a matting result prediction block after the upsample block. Each prediction block contains a convolution layer, batch normalization layer, ReLU and convolution layer sequentially.

\noindent \textbf{Progressive Refinement Module.} 
We adopt Progressive Refinement Module (PRM)~\cite{yu2021mask} to further refine the output matting alphas from output blocks. PRM can selectively fuse the matting alphas from the previous scale and the current scale with a self-guidance mask, which can preserve the confident regions from the previous scale and focus on refining uncertain regions at the current scale. Specifically, the self-guidance mask of the current scale is generated from matting alpha obtained at the previous scale as follows:
\begin{equation}
    g_l = \begin{cases}
0& \text{$0~<~\alpha_{l-1}~<~1$},\\
1& \text{otherwise},
\end{cases}
\label{gl}
\end{equation}
where $\alpha_{l-1}$ is the matting alpha of previous scale.The $\alpha_{l-1}$ is upsampled to match the size of the raw matting output $\alpha'_{l}$ of the current scale. With the self-guidance mask $g_l$, the refined matting alpha of current scale can be calculated as following:
\begin{equation}
    \alpha_l = \alpha'_lg_l+\alpha_{l-1}(1-g_l).
\end{equation}

Note that features in confident region predicted from the previous scale are preserved and the current scale only focuses on refining the uncertain region.

\section{Training}
We train our SGHM model in two stages. First, we train the segmentation network using widely available segmentation datasets. In this stage, the parameters of shared encoder and segmentation decoder are updated simultaneously. After segmentation net is trained, the shared encoder can extract powerful semantic features to provide information for the matting task. Next, we fix the shared encoder and segmentation decoder, and train the matting decoder only. The coarse mask output from segmentation net is also used as the input at this stage. During inference, the two decoders are executed successively. Segmentation mask is first predicted and fed to matting decoder to produce matting result.

\subsection{Segmentation Training}
We train the segmentation sub-network with about $47.2k$ paired images, which are from SPD\cite{spd} (about $2.5k$), Portrait Matting~\cite{shen2016deep} (about $1.7k$), dataset released in ~\cite{wu2014early}(about $5.2k$), human parsing dataset~\cite{gong2018instance} (about $4.7k$), Privacy-Preserving dataset~\cite{li2021privacy} (about $9.4k$) and green screen dataset from BMV2~\cite{bgmv2} (about $23.7k$). We treat green screen data as segmentation mask since it provides more body posture diversity than alpha details. Note that we drop some image pairs by annotation checking, and collect about $35k$ background images from internet for random background composition.

We adopt Binary Cross Entropy (BCE) loss to train segmentation model. For data augmentation, we adopt random affine transformation, random horizontal flipping, random noise, random color jitters, random composite, and random crop to $320\times320$. We train our segmentation model on 8 NVIDIA Tesla A100 GPUs with a batch size of 10 for each GPU. We use Adam as optimizer, and the learning rate is initialized to $5e^{-4}$. The model is totally trained for 100 epochs with a cosine learning rate decay scheduler. 

\subsection{Matting Training}
We train matting model on the foreground images of AIM~\cite{deepimagematting} dataset except transparent object images. The total foreground images are 269, and we use MS COCO dataset as background images. 

Following MG~\cite{yu2021mask}, we adopt $l_1$ regression loss, composition loss~\cite{deepimagematting}, Laplacian loss~\cite{Hou_2019_ICCV} for training matting model. We denote the ground-truth alpha with $\hat{\alpha}$ and the prediction alpha with $\alpha$. Then the combined loss function can be formulated as:
\begin{equation}
L(\hat{\alpha},\alpha) = L_{l_1}(\hat{\alpha},\alpha) + L_{comp}(\hat{\alpha},\alpha)+ L_{lap}(\hat{\alpha},\alpha).
\end{equation}

We apply this combined loss on all output matting alphas at $1, \frac{1}{4},\frac{1}{8}$ scale with adaptive weights $g_l$ calculated in Eq.~\ref{gl} to force the training to more focused on the unknown region at each scale. Moreover, we set different weights for different scales to form the final loss function as follows:
\begin{equation}
    L_{tot} = \sum_l{\omega_{l}L(\hat{\alpha}_{l} \cdot g_{l},\alpha_{l}\cdot g_{l})},
\end{equation}
where $\omega_{l}$ is the loss weight of different scales. We set $\omega_{\frac{1}{8}}:\omega_{\frac{1}{4}}:\omega_{1}=1:2:3$ in our experiments. 

We train our matting model with $100,000$ iterations on 4 NVIDIA Tesla A100 GPUs with a batch size of 8 for each GPU. We use Adam as optimizer, and the learning rate is initialized to $1e^{-3}$. We adopt the same data augmentation with training of segmentation, with a random crop of $1280\times1280$. We also adopt mask perturbation for augmentation. Note that we fix the parameters of segmentation model during training matting module, which can force the matting decoder to focus more on features to predict alpha details. If not fixed matting performance will drop as it will overfit to the small set of matting data.

\section{Experiment}
\subsection{Benchmarks}
To verify the effectiveness of the proposed method, we evaluate the performance on the following $5$ benchmarks, including three real-world datasets and two composition datasets.

\noindent \textbf{AIM~\cite{deepimagematting}.} We select $12$ human images from AIM dataset for testing. Each foreground human image is composited to 20 backgrounds which are selected from top-240 of BG-20K\cite{li2020end} test set.

\noindent \textbf{D646~\cite{qiao2020attention}.} Similar to AIM, $11$ foreground images are composited with the last 220 backgrounds from BG-20K test set.

\noindent \textbf{PPM-100~\cite{MODNet}.} This dataset provides 100 finely annotated portrait images with various backgrounds. Images from PPM-100 are more realistic and natural than composition images.

\noindent \textbf{P3M-500-NP~\cite{li2021privacy}.} We use the face kept images rather than face masked from P3M. The purpose is to avoid the unknown impact of face blur on evaluation. This benchmark has a great diversity of body postures.

\noindent \textbf{RWCSM-289.} To further verify our model generalization, we build a real-world complex scene matting dataset, denoted as RWCSM-289. It contains a variety of complex living and working scenarios. Its sources are hand-hold captured videos, online video meetings, TV shows, live videos, and Vlogs. Many of them come from youtube and are used by RVM~\cite{lin2022robust}. It is worth noting that this dataset include motion and multi-person scenes, which is helpful to evaluate model robustness. The ground truth alpha is annotated by PhotoShop.

\begin{table}[ht]
\setlength{\tabcolsep}{8pt}
\begin{center}
\caption{Quantitative results on real-world benchmarks. {'$\downarrow$'} : lower values are better.}
\label{tab:realworld}
\begin{tabular}{cccccc}
\hline\noalign{\smallskip}
Dataset  & Method & MAD{$\downarrow$} & MSE{$\downarrow$} & Grad{$\downarrow$} & Conn{$\downarrow$} \\
\noalign{\smallskip}\hline\noalign{\smallskip}
& LFM& 15.80&9.40&-&-\\
& SHM& 15.20&7.20&-&-\\
& HATT& 13.70&6.70&-&-\\
PPM-100 & BSHM& 11.40&6.30&-&-\\
& P3MNet& 15.61&12.86&56.37&130.42\\
& MODNet& 8.60&4.40&64.26&80.82\\
& RVM&10.95&6.53&63.13&105.19\\
& SGHM (ours)&\textbf{5.97}&\textbf{2.58}&\textbf{48.20}&\textbf{51.17}\\
\noalign{\smallskip}\hline\noalign{\smallskip}
& LFM&18.80& 13.10&31.93&19.50\\
& SHM&12.20&9.30&20.30&17.09\\
& HATT&17.60&7.20&19.99&27.42\\
P3M-500-NP &P3MNet& 6.50&3.50&\textbf{10.35}&12.51\\
& MODNet&12.82&7.41&16.02&20.23\\
& RVM&11.10&7.06 &15.30 &19.17 \\
& SGHM (ours)&\textbf{6.49}&\textbf{3.11}&11.39&\textbf{10.16}\\
\noalign{\smallskip}\hline\noalign{\smallskip}
&P3MNet&32.92&31.09&28.42&77.37\\
RWCSM-289 &MODNet&18.95&15.76&19.65&46.18\\
& RVM&14.36&11.25&15.68&28.52\\
& SGHM (ours) &\textbf{9.23}&\textbf{6.57}&\textbf{13.52}&\textbf{18.68}\\
\noalign{\smallskip}\hline
\end{tabular}
\end{center}
\end{table}

\begin{figure*}[htb]
	\centering
	\includegraphics[width=\linewidth]{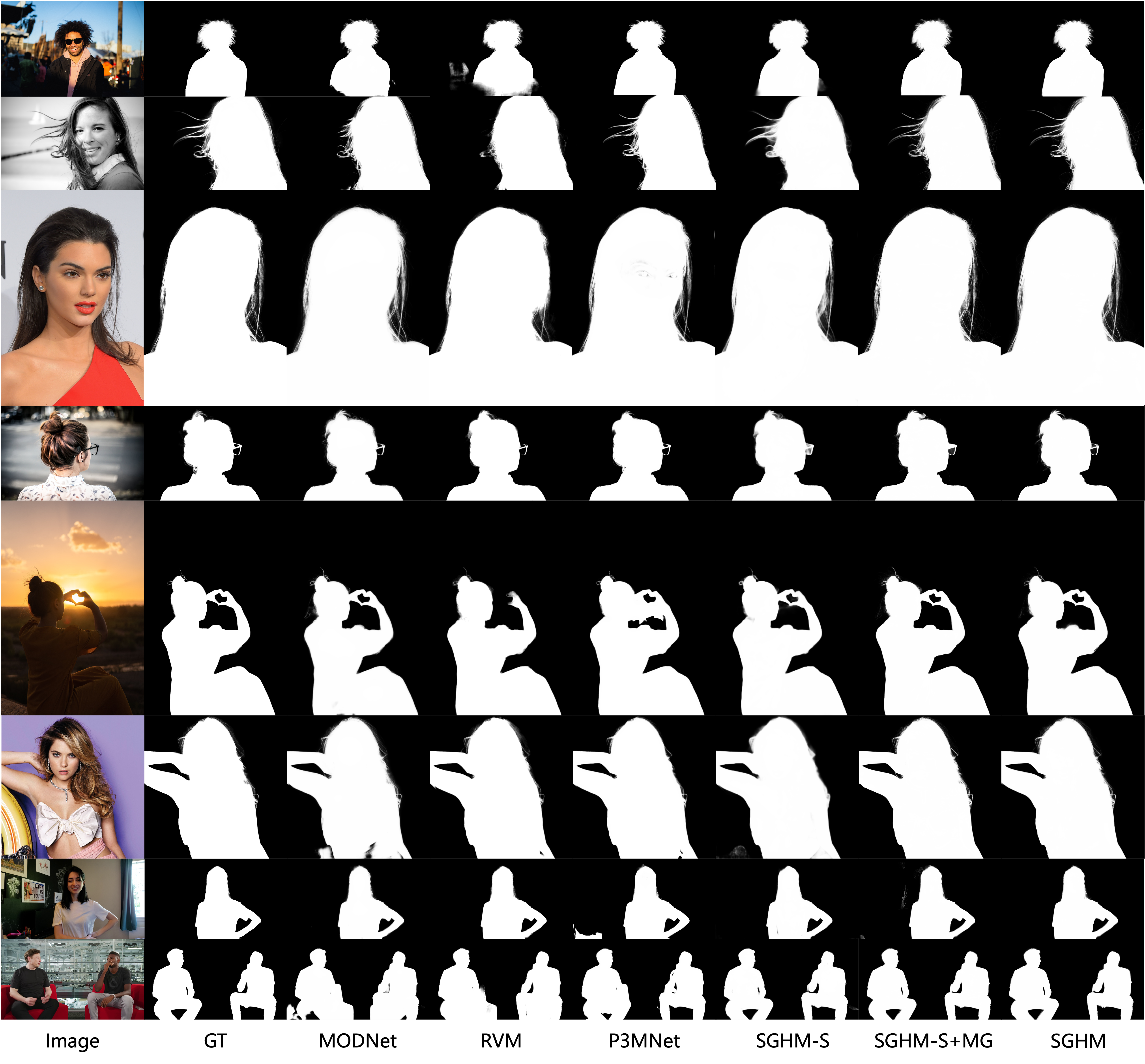}
	\caption{{Visual comparison of different methods on alpha details. SGHM-S denotes the segmentation results of our method. SGHM denotes the final matting results. SGHM-S+MG denotes using SGHM-S as extra input for MG. Our proposed method produces superior results from coarse to fine. Best viewed on monitor with zooming in for detail.}}
	\label{fig:results_alpha}
\end{figure*}

\begin{table}[ht]
\setlength{\tabcolsep}{10pt}
\begin{center}
\caption{Quantitative results on composition benchmarks. {'$\downarrow$'} : lower values are better.}
\label{tab:composition}
\begin{tabular}{cccccc}
\hline\noalign{\smallskip}
Dataset  & Method & MAD{$\downarrow$} & MSE{$\downarrow$} & Grad{$\downarrow$} & Conn{$\downarrow$} \\
\noalign{\smallskip}\hline\noalign{\smallskip}
&P3MNet&44.78&37.70&43.02&100.80\\
AIM&MODNet&33.18&23.58&29.08&74.47\\
& RVM&27.07&17.54&28.84&60.73\\
& SGHM (ours) &\textbf{14.34}&\textbf{7.18}&\textbf{19.29}&\textbf{29.40}\\
\noalign{\smallskip}\hline\noalign{\smallskip}
&P3MNet&20.25&15.27&36.93&54.74\\
D646&MODNet&10.52&4.72&32.62&28.61\\
& RVM&10.50&4.94&35.24&28.60\\
& SGHM (ours) &\textbf{6.59}&\textbf{2.19}&\textbf{19.07}&\textbf{17.02}\\
\noalign{\smallskip}\hline
\end{tabular}
\end{center}
\end{table}

\setlength{\tabcolsep}{11pt}
\begin{table}[tb]
\begin{center}
\caption{Size and Speed Comparison. The matting metrics are evaluated on PPM-100 dataset and the speed is evaluated on HD size on an NVIDIA A100 GPU. SGHM-S is the segmentation branch of our method. It runs at over 100 FPS as it is on 1/4 of full image resolution.}
\label{table:speed}
\begin{tabular}{lcccc}
\hline\noalign{\smallskip}
Method  & \#Parameters (M) & FPS & MAD & MSE\\
\noalign{\smallskip}\hline\noalign{\smallskip}
MODNet&6.49&20.76&8.60&4.40\\
RVM&3.75&71.81&10.95&6.53\\
SGHM-S&40.22&106.14&11.84&5.72\\
SGHM-S+MG&69.92&18.14&8.83&4.18\\
SGHM&43.94&34.76&5.97&2.58\\
\noalign{\smallskip}\hline
\end{tabular}
\end{center}
\end{table}

\subsection{Quantitative Comparison}
We compare our approach with the state-of-the-art automatic matting methods, including LFM~\cite{zhang2019late}, SHM~\cite{chen2018semantic}, HATT~\cite{qiao2020attention}, BSHM~\cite{liu2020boosting}, MODNet~\cite{MODNet}, P3MNet~\cite{li2021privacy}, video matting method RVM~\cite{lin2022robust} and mask-guided method MG~\cite{yu2021mask}. We use inference size 512 for MODNet since it provides the best results on PPM-100. For RVM, We generate 10 frames video by repeating 10 times for every single image and take last frame result as evaluation target. For P3MNet the recommended testing resize strategy is used. For MG, we feed our segmentation result to its network as mask guidance. Both MG and our method keep the short size of images to 1280 when testing. We use mean absolute difference (MAD), mean squared error (MSE), spatial gradient  (Grad)~\cite{rhemann2009perceptually}, and connectivity (Conn)~\cite{rhemann2009perceptually} as alpha matting quality metrics. Note that MAD and MSE values are scaled by $10^{3}$ and all metrics are calculated over the whole image.

Table~\ref{tab:realworld} and Table~\ref{tab:composition}  show the results of different matting methods evaluated on real-world and composition datasets. It shows that our method outperforms other methods across all real-world datasets in all metrics. Specifically, our method is ahead of compared method on PPM-100. On P3M-500-NP, we achieve the results (MAD $6.49$, MSE $3.11$) that are on par with the P3MNet (MAD $6.50$, MSE $3.50$) by only introducing face-masked P3M data into the segmentation stage. For complex scene data RWCSM-289 which covers more diversity of background, number of humans, body accessories, illumination, and image resolution, we significantly outperform P3MNet and MODNet, and are better than video-based approach RVM. On the composition datasets, SGHM still achieves the best results, showing consistently excellent performance of the proposed method.

\subsection{Qualitative Comparison}

\begin{figure}[tb]
\centering
\includegraphics[height=10.0cm]{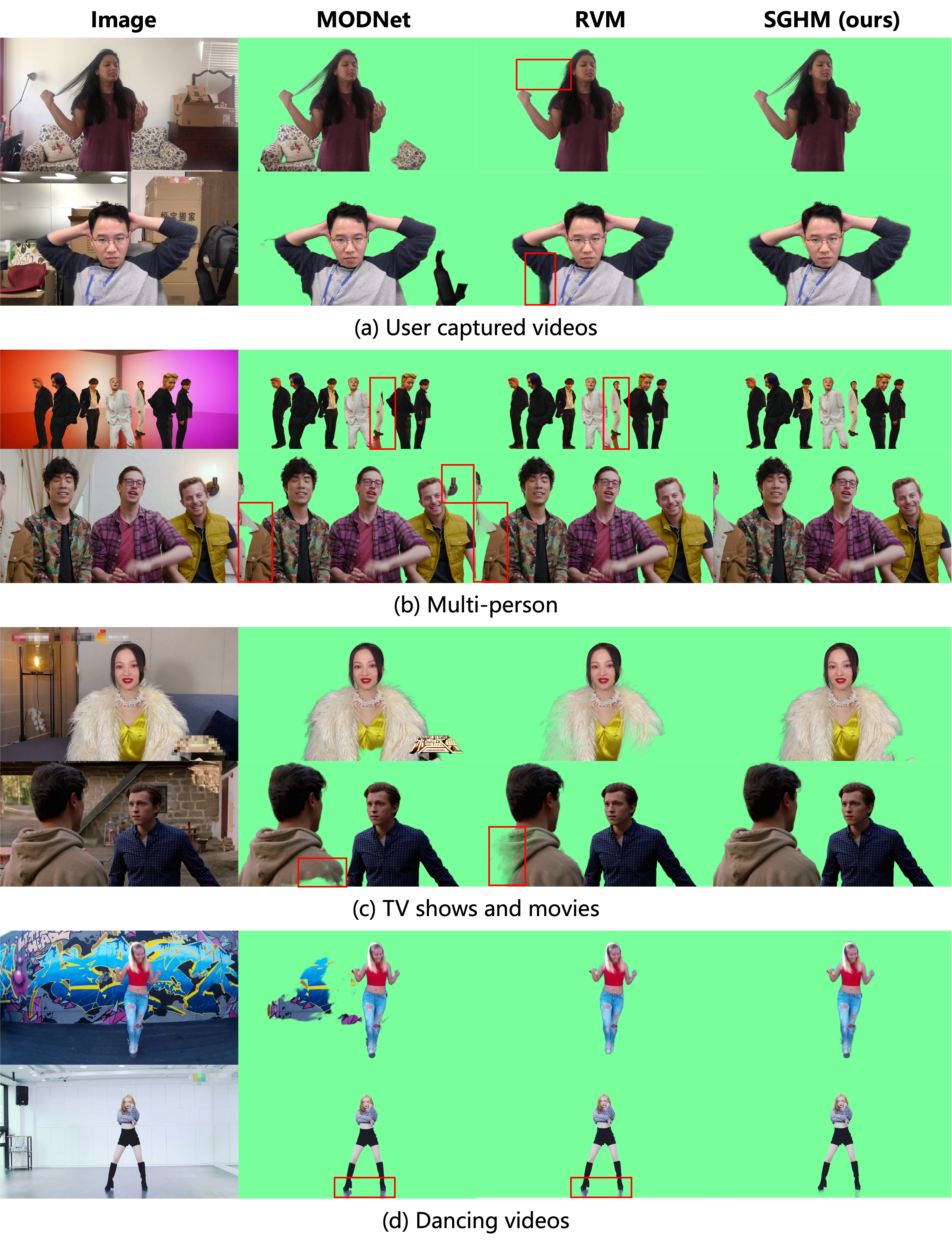}
\caption{Visual comparison of different methods on four categories of videos. Our method is more robust to semantic errors.}
\label{fig:results_green}
\end{figure}

This section shows qualitative comparisons on real-world benchmarks. We reveal alpha details in Fig.~\ref{fig:results_alpha} and model robustness in Fig.~\ref{fig:results_green}. In Fig.~\ref{fig:results_alpha} rows 1 to 4, we compare hair details and find ours predict fine-grained hair  details comparable to mask-based method MG, which are more accurate than P3MNet, MODNet and RVM. Multiple body postures are displayed in rows 5 to 8. Other methods tend to get semantic errors (can be found in MODNet at row of 6, MG at row of 7, P3MNet at row 5) while SGHM produces more accurate alpha matte. It is worth noting that our method has the ability to correct semantic errors in the coarse masks (see row 1, 6 and 8 from SGHM-S to SGHM). 

In Fig.~\ref{fig:results_green}, we select two SOTA methods MODNet and RVM for robustness comparison from four categories of videos. The extracted foreground is composited with a green background for visualization. Our method predicts much fewer semantic errors and demonstrates better robustness against semantic understanding errors than the other two methods.

\subsection{Size and Speed Comparison}
As mentioned in Section 3, MG uses a segmentation mask as extra input to its matting network. Unlike MG, we incorporate the mask generation stage into matting framework. Since MG uses an independent matting network, it introduces more parameters and its total parameter number is the combination of segmentation and matting networks. The speed is thus slowed down. Our SGHM shares the encoder with the segmentation, which causes marginal extra parameters. SGHM also runs faster than MG on the same setting and can achieve 34 FPS on HD image ($1920 \times 1080$) on NVIDIA A100. For the matting quality, our method achieves better performance. This shows we have both speed and accuracy advantages over MG. MODNet and RVM are also compared. Although they have fewer parameters, they both have limitation. MODNet runs slower on HD inference size (20.76 FPS) than 512 (81.01 FPS). RVM predicts unsatisfactory fine-grained alpha results across all benchmarks.

\setlength{\tabcolsep}{8pt}
\begin{table}[t]
\begin{center}
\caption{Ablation study on different settings, tested on PPM-100.}
\label{tab:ablation}
\begin{tabular}{ccc|cc}
\hline\noalign{\smallskip}
ASM & Mask guidance & Sharing  encoder weights & MAD & MSE   \\
\noalign{\smallskip}\hline\noalign{\smallskip}
$\surd$ &  &  & 17.23 & 8.54\\
 & & $\surd$ & 10.46 & 5.07\\
$\surd$ & $\surd$ &  & 7.83 & 4.04\\
 & $\surd$ & $\surd$ & 7.50 & 3.52\\
$\surd$ & $\surd$ & $\surd$ & \textbf{5.97} & \textbf{2.58}\\
\noalign{\smallskip}\hline
\end{tabular}
\end{center}
\end{table}

\subsection{Ablation Studies}
\noindent \textbf{Role of Segmentation Task.} We propose to introduce segmentation task to improve the performance and generalization of alpha matting in two ways. One is sharing encoder features and other is coarse mask guidance. Table~\ref{tab:ablation} shows our ablation study results on PPM-100. The results lead to two conclusions: (1) Sharing semantic features is very beneficial to matting task, which helps to reduce MAD from 7.83 to 5.97. (2) Mask guidance plays an indispensable role in guiding matting process, as matting performance drops dramatically when mask guidance is  removed. Since our matting model is trained on only hundreds of images, robust semantic features and good mask guidance are both helpful for improving model generalization. 

\setlength{\tabcolsep}{11pt}
\begin{table}[ht]
\begin{center}
\caption{Results of different training datasets sizes, tested on PPM-100. The LargeSeg dataset consists of 140k human masks which are collected from multiple publicly available datasets. The size of D646 is 362 which is selected from the Distinctions-646 training set.}
\label{tab:training size}
\begin{tabular}{lcccc}
\hline\noalign{\smallskip}
 & Segmentation&Matting &  &   \\
 & Dataset Size & Dataset Size & MAD & MSE  \\
\noalign{\smallskip}\hline\noalign{\smallskip}
Baseline &40k&200+&5.97&2.58\\
 +LargeSeg &170k&200+&\textbf{5.16}& \textbf{2.04}\\
 +D646&40k &600+&5.71&2.45\\
\noalign{\smallskip}\hline
\end{tabular}
\end{center}
\end{table}
\setlength{\tabcolsep}{1.4pt}

\noindent \textbf{Role of ASM.} We propose ASM to combine semantic features and segmentation mask for matting decoder. As listed in the fourth and fifth rows of Table~\ref{tab:ablation}, model gets worse results without ASM. SpectralNorm and ECA layer are the two key components in ASM. In-depth analysis reveals that MAD drops from 5.97 to 7.11 when ECA layer is removed, while MAD is 6.33 when SpectralNorm is removed. This indicates ECA layer contributes more as it channel-wisely re-weight the features to adapt them for matting. Note that in the first row, we remove the mask input to only use the features from encoder and keep the same Conv layers with proposed ASM.

\noindent \textbf{Role of Dataset Size.}
We further conduct an experiment to verify the data efficiency of our method. As can be seen in Table~\ref{tab:training size}, a larger segmentation dataset improves matting results significantly, while increasing the matting dataset size improves slightly. Note that it is easy to collect these human segmentation masks from publicly available datasets. But labeling fine-grained matting requires a much higher annotation skill level and it is time and money costing. This is an important and practical finding that we can efficiently improve matting performance by collecting more coarse human masks in an easy and fast way rather than paying for the high cost fine-detailed alpha annotating.

\section{Conclusion}
In this work, we investigate the major challenge in robust human matting and reveal that it is from the semantic understanding. Based on this, we propose a semantic guided human matting method. We introduce an additional matting decoder to the semantic segmentation network. By reusing the features from semantic segmentation encoder, the matting decoder is aware of global semantic information and also can generate fine matting details. With very small number of matting data, we can train a robust, accurate and real-time  matting model which achieves top performance on multiple benchmark datasets. We believe that our proposed framework is a practical pipeline for matting application which does not rely on large number of high annotation cost matting data.

%
%
%

\end{document}